\documentclass[10pt,conference]{IEEEtran}
\usepackage{xcolor,soul,framed} 
\usepackage[noadjust]{cite}

\colorlet{shadecolor}{yellow}
\usepackage[bookmarks=false]{hyperref}
\usepackage[pdftex]{graphicx}
\graphicspath{{../pdf/}{../jpeg/}}
\DeclareGraphicsExtensions{.pdf,.jpeg,.png}
\usepackage[caption=false, font=footnotesize]{subfig}
\usepackage{amssymb,mathtools}
\usepackage{array}
\usepackage{pifont}
\usepackage{mdwmath}
\usepackage{mdwtab}
\usepackage{eqparbox}
\usepackage{url}
\usepackage{tikz}
\usetikzlibrary{shapes.geometric,arrows.meta,positioning,fit,backgrounds,calc}
\usetikzlibrary{decorations.pathreplacing}
\usepackage{enumitem} % for left margin
\usepackage[mathcal]{euscript} % follow IEEE proofreading
\usepackage{tabularx} % for table of equations
\usepackage{multirow} % for multiple sub-rows of a cell of a table
\usepackage{booktabs} % for table
\usepackage{makecell} % for thick hline
\usepackage{aligned-overset} % for align overset
\usepackage{url}
\usepackage{multicol} % for two equations side-by-side
\usepackage{algorithm}
\usepackage{algorithmicx}
\usepackage{algpseudocode}
\usepackage{pgfplots}
\pgfplotsset{compat=1.18}
\usepackage{balance}
\usepackage{bbm}
\usepackage{amsmath}
\usepackage{amsthm} % for bold Lemma, Theorem, etc., with italic text
\usepackage{pifont}
\newcommand{\cmark}{\textcolor{green!80!black}{\ding{51}}}
\newcommand{\xmark}{\textcolor{red}{\ding{53}}}

\graphicspath{{./Figures/}}

\definecolor{grpA}{RGB}{220,50,47}
\definecolor{grpB}{RGB}{38,139,210}
\definecolor{grpC}{RGB}{42,161,152}
\definecolor{grpD}{RGB}{181,137,0}
\definecolor{stageblue}{RGB}{52,101,164}
\definecolor{stagegray}{RGB}{136,138,133}

\tikzset{
  pipe/.style={draw=stageblue,thick,-{Stealth[length=5pt]}},
  stagebox/.style={rectangle,rounded corners=4pt,draw=stageblue,
                   fill=stageblue!12,text width=2.1cm,align=center,
                   minimum height=0.75cm,font=\small\bfseries},
  kgnode/.style={circle,draw,thick,fill=white,
                 minimum size=0.7cm,font=\scriptsize\bfseries,align=center},
  kgedge/.style={draw,-{Stealth[length=4pt]},thick},
}

\begin{document}
      \title{Knowledge-Guided LLM for Intent-Driven Service Orchestration in Integrated Terrestrial and Non-Terrestrial 6G Networks}
\author{
  \IEEEauthorblockN{Wajdi Hammami, Irfan Azam, and Soumaya Cherkaoui}
  \IEEEauthorblockA{Department of Computer and Software Engineering, Polytechnique Montreal, Canada} \IEEEauthorblockA{Email: \{\texttt{wajdi.hammami, irfan.azam, soumaya.cherkaoui}\}@polymtl.ca}
}

\maketitle
\begin{abstract}
Natural-language service intents in integrated terrestrial and non-terrestrial (TN/NTN) 6G networks can be incomplete, ambiguous, or inconsistent, making unconstrained LLM generation challenging for reliable network-facing orchestration. This paper proposes a knowledge-guided LLM framework that combines grammar-constrained semantic parsing with a knowledge-graph (KG)-driven clarification loop to produce validated, orchestration-ready service profiles across four representative service classes. On a 200-sample intent dataset, the framework achieves \(89\%\) classification accuracy and \(95\%\) policy validity with an average of \(0.66\) clarification turns per request. Compared with a Pure-LLM baseline, the proposed framework improves policy validity and contradiction detection while requiring fewer clarification turns, demonstrating the value of explicit knowledge-guided reasoning for robust intent processing in integrated TN/NTN 6G orchestration.
\end{abstract}

% === I. INTRODUCTION 
%%%%%%%%%%%%%%%%%%%%%%%%%%%%%%%%%%%%%%%%%%%%%%%%%%%%%%%%%%%%%%%%%%%%%%%%%%%%%%%%
\section{INTRODUCTION}\label{sec:intro}
Sixth generation (6G) wireless systems are envisioned to evolve beyond incremental improvements in throughput and latency toward a deeply integrated, intelligent, and service-aware communication fabric capable of supporting ubiquitous connectivity across diverse environments and application domains. The integration of terrestrial and non-terrestrial segments has emerged as a key architectural direction for extending coverage, improving resilience, and enabling seamless service continuity in remote, rural, maritime, aerial, and disaster-affected scenarios \cite{liu2018sagin}. By combining terrestrial infrastructure with satellites, aerial platforms, and other non-terrestrial assets, integrated terrestrial/non-terrestrial networks (TN/NTN) are expected to become a fundamental pillar of future 6G ecosystems \cite{wang2021maritime}. 

At the same time, the coexistence of heterogeneous platforms, dynamic topologies, and multi-domain operational constraints makes traditional static network management increasingly inadequate. Future 6G systems are therefore expected to operate under intent-driven and AI-native principles, where users or operators express objectives at a high level and the network interprets them into actionable decisions \cite{boutouchent20256gintense,meng2024semantic}. In integrated TN/NTN settings, such decisions must account for resilience requirements, fallback preferences, latency sensitivity, and heterogeneous infrastructure conditions making the intent interpretation problem substantially more complex than in single-domain settings.

Large language models (LLMs) have recently emerged as a promising tool for semantic interaction and high-level intent understanding in communication networks \cite{manias2024semantic,mekrache2025ossgpt}. In parallel, semantic and AI-native architectures for integrated TN/NTN environments further motivate the need for intelligent intent processing at higher network layers \cite{qiu2026gost}. However, in hybrid TN/NTN environments, service orchestration cannot rely on unconstrained LLM generation alone. Natural-language service requests can be incomplete, ambiguous, or internally inconsistent, while orchestration decisions must remain structured, reliable, and compliant with domain-specific service constraints. This creates a need for hybrid frameworks in which LLMs provide semantic understanding and a knowledge-guided layer ensures validity, consistency, and network relevance.

\begin{table*}[t]
\centering
\caption{Feature-wise comparison with representative related works. \cmark: addressed; \xmark: not addressed; $\sim$: partially addressed.}
\label{tab:feature_comparison}
\footnotesize
\renewcommand{\arraystretch}{1.02}
\setlength{\tabcolsep}{4pt}
\begin{tabular}{>{\raggedright\arraybackslash}p{4.2cm}|>{\centering\arraybackslash}p{1.1cm}|>{\centering\arraybackslash}p{1.5cm}|>{\centering\arraybackslash}p{1.7cm}|>{\centering\arraybackslash}p{1.6cm}|>{\centering\arraybackslash}p{1.4cm}|>{\centering\arraybackslash}p{2.0cm}}
\toprule
\textbf{Representative Work} & \textbf{LLM} & \textbf{Knowledge Graph} & \textbf{Multi-Turn Clarification} & \textbf{Constrained Generation} & \textbf{Integrated TN/NTN} & \textbf{Structured Orches. Output} \\
\midrule
Meng \emph{et al.} \cite{meng2024semantic}       & \cmark & \cmark & \xmark & \xmark & \cmark & \xmark \\
Manias \emph{et al.} \cite{manias2024semantic}   & \cmark & \xmark & \xmark & \xmark & \xmark & \xmark \\
Mekrache \emph{et al.} \cite{mekrache2025ossgpt} & \cmark & \xmark & $\sim$  & \xmark & \xmark & \cmark \\
Qiu \emph{et al.} \cite{qiu2026gost}             & \xmark & \cmark & \xmark & \xmark & \cmark & \xmark \\
Tang \emph{et al.} \cite{tang2025airespo}        & \cmark & \xmark & \xmark & \xmark & \xmark & \cmark \\
Habib \emph{et al.} \cite{habib2025ran}          & \cmark & \xmark & \xmark & \xmark & \xmark & \xmark \\
Leivadeas \emph{et al.} \cite{leivadeas2023ibn}  & \xmark & \xmark & \xmark & \xmark & \xmark & \xmark \\
Mondal \emph{et al.} \cite{hotnets2025ambiguity} & \cmark & \xmark & \cmark & \xmark & \xmark & \xmark \\
Mehmood \emph{et al.} \cite{mehmood2024KG}       & \xmark & \cmark & \xmark & \xmark & \xmark & \xmark \\
\midrule
\textbf{Proposed}                                & \cmark & \cmark & \cmark & \cmark & \cmark & \cmark \\
\bottomrule
\end{tabular}
\end{table*}

Recent works span LLM-assisted orchestration, intent-based network management, and structured reasoning for network automation. LLM-assisted systems have shown practical value in 5G core management, operation support, AI-RAN service repositories \cite{tang2025airespo}, and intent-driven RAN management \cite{habib2025ran}, but rely on unconstrained or weakly constrained generation and do not explicitly address TN/NTN-specific attributes such as deployment mode and fallback behavior. Intent-based networking (IBN) has also been studied as a paradigm for high-level network management \cite{leivadeas2023ibn}, formalized through standards such as 3GPP TS~28.312 \cite{3gpp28312}, and extended toward next-generation and LLM-enabled intent management \cite{ayoubi2024ibn,netintent2025,phone2026}. Related knowledge-driven efforts have explored ambiguity in LLM-based network configuration \cite{hotnets2025ambiguity} and knowledge graph embedding for intent processing and validation in IBN pipelines \cite{mehmood2024KG}, but these works do not jointly address ambiguity resolution, contradiction detection, and deployment-aware service-profile generation in an integrated TN/NTN orchestration setting \cite{latestNTN2026}. Grammar-constrained decoding has been shown to enforce structural validity at the token level \cite{willard2023constrained,xgrammar2025,crane2025}, while knowledge-guided approaches show that grounding LLM reasoning in structured domain knowledge improves faithfulness and reduces hallucination \cite{ograg2025,graphrag2025}. However, no prior work jointly combines KG-guided dialogue management, grammar-constrained structured generation, and integrated TN/NTN-aware service orchestration within a unified intent-driven framework. Table~\ref{tab:feature_comparison} summarizes the position of the proposed framework relative to representative prior works across the main capabilities relevant to intent-driven TN/NTN service orchestration.

\par\noindent\textbf{Motivation:} Despite these advances, an important gap remains. To the best of our knowledge, no existing framework jointly accepts natural-language service intents in integrated TN/NTN 6G environments, resolves ambiguity and inconsistency through knowledge-guided validation and clarification, and produces orchestration-ready service profiles meaningful for integrated TN/NTN service support. Unconstrained prompting alone may yield incomplete or inconsistent outputs for network-facing use, while a purely rule-based process loses the flexibility and semantic richness of intent-driven interaction; a robust hybrid approach is therefore required.

\par\noindent\textbf{Contributions:} To address this gap, we propose a knowledge-guided LLM framework for intent-driven service orchestration in integrated TN/NTN 6G networks, combining an LLM semantic understanding layer, a knowledge-guided reasoning and validation layer that enforces service constraints and supports intent clarification, and a service orchestration layer that maps the validated intent into a structured orchestration-ready service profile.

The main contributions of this paper are as follows:
\begin{itemize}[itemsep=1pt,topsep=2pt]
    \item We formulate the problem of intent-driven service orchestration in integrated TN/NTN 6G networks, where natural-language service requests must be converted into structured and network-relevant orchestration decisions.
    \item We propose a knowledge-guided LLM framework in which the LLM provides semantic understanding of service intents, while a domain knowledge layer constrains reasoning, detects missing or conflicting requirements, and validates the generated service representation.
    \item We develop an orchestration-oriented service decision mechanism that maps service intents into structured profiles for integrated TN/NTN service support, including service class, deployment preference, fallback behavior, and policy-relevant attributes.
    \item We establish an intent-level evaluation framework for assessing classification accuracy, policy validity, contradiction handling, and clarification efficiency in knowledge-guided TN/NTN service orchestration.
\end{itemize}

%%%%%%%%%%%%%%%%%%%%%%%%%%%%%%%%%%%%%%%%%%%%%%%%%%%%%%%%%%%%%%%%%%%%%%%%%%%%%%%%
% === III. SYSTEM MODEL AND PROBLEM FORMULATION
%%%%%%%%%%%%%%%%%%%%%%%%%%%%%%%%%%%%%%%%%%%%%%%%%%%%%%%%%%%%%%%%%%%%%%%%%%%%%%%%
\vspace{-1em}
\section{SYSTEM MODEL AND PROBLEM STATEMENT}
\label{sec:system_model}

We consider an intent-driven orchestration framework for integrated TN/NTN 6G networks. A user, operator, or vertical application submits a natural-language service request, and the objective is to transform it into a validated, orchestration-ready service profile suitable for TN-only, NTN-only, or hybrid TN/NTN support.
\vspace{-1em}
\subsection{System Architecture}
The architecture involves three main entities:
\begin{itemize}[itemsep=1pt,topsep=2pt]
    \item a service requester that provides a natural-language intent,
    \item a knowledge-guided LLM orchestration engine that interprets, validates, clarifies, and structures the request,
    \item an integrated TN/NTN network context that provides the deployment conditions and scenario constraints for service support.
\end{itemize}

The input intent $\mathbf{i} \in \mathcal{I}$, where $\mathcal{I}$ denotes the space of admissible natural-language service requests, may be incomplete, ambiguous, or inconsistent and therefore cannot be directly consumed by downstream orchestration functions. The orchestration engine maps $\mathbf{i}$ to a structured service profile $\mathcal{P} \in \mathcal{P}_{\mathrm{valid}}$, the set of ontology-compliant profiles, which is then used to determine deployment mode, fallback behavior, and service-support policy.
\vspace{-1em}
\subsection{Input and Output Representation}
The target output of the framework is the orchestration-ready profile
\begin{equation}
\mathcal{P} =
\left\{
c,\,
e,\,
\ell,\,
r,\,
\tau,\,
\gamma,\,
p,\,
f,\,
\kappa,\,
m,\,
d
\right\},
\label{eq:profile}
\end{equation}
where $c \in \mathcal{C}$ is the service class, $e \in \mathcal{E}$ is the environment type, $\ell \in \mathcal{L}$ is the latency class, $r \in \mathcal{R}$ is the reliability class, $\tau \in \mathcal{T}$ is the throughput class, $\gamma \in \Gamma$ is the coverage scope, $p \in \mathcal{P}_{r}$ is the priority level, $f \in \mathcal{F}_{b}$ is the fallback requirement, $\kappa \in \mathcal{K}_{c}$ is the cost sensitivity, $m \in \mathcal{M}$ is the mobility level, and $d$ is the deployment mode drawn from $\mathcal{D}=\{\mathrm{TN\text{-}only},\,\mathrm{NTN\text{-}only},\,\mathrm{Hybrid}\}$.

\subsection{Representative Service Classes}
To keep the framework representative yet tractable, we consider four service classes: Time Critical (TC), Mission Critical (MC), Remote Connectivity (RC), and Massive Connectivity (MaC), summarized in Table~\ref{tab:service_classes}. These classes define the top-level service taxonomy used for intent classification, validation, and orchestration.

\begin{table}[t]
\centering
\caption{Representative service classes in the proposed framework.}
\label{tab:service_classes}
\footnotesize
\renewcommand{\arraystretch}{1.02}
\setlength{\tabcolsep}{3.5pt}
\begin{tabular}{c|l|l}
\toprule
\textbf{Class} & \textbf{Primary Objective} & \textbf{Likely Mode} \\
\midrule
TC  & Stable low-latency service support & TN / Hybrid \\
MC  & Resilient and continuous service support & Hybrid / NTN \\
RC  & Wide-area service reach in sparse/degraded regions & Hybrid / NTN \\
MaC & Scalable support for sensing and telemetry services & NTN / Hybrid \\
\bottomrule
\end{tabular}
\end{table}

Specifically, TC emphasizes low-latency control, MC emphasizes resilience and continuity, RC emphasizes wide-area service reach, and MaC emphasizes scalable sensing and telemetry support.

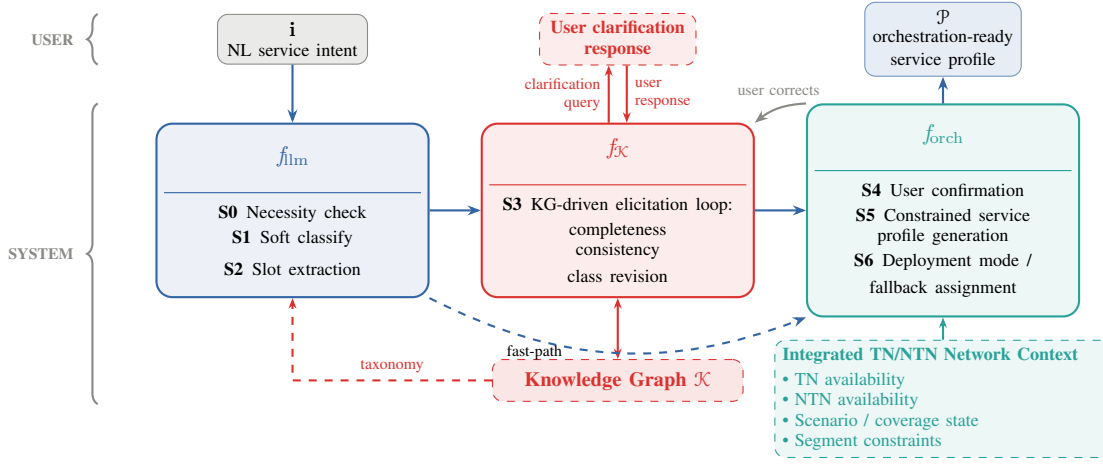
\begin{figure*}[!t]
\centering
\begin{tikzpicture}[
  node distance=0.8cm,
  blkllm/.style={rectangle, rounded corners=6pt, draw=stageblue, thick,
                 fill=stageblue!9, text width=4.0cm, align=center,
                 minimum height=2.7cm},
  blkkg/.style={rectangle, rounded corners=6pt, draw=grpA, thick,
                fill=grpA!9, text width=4.0cm, align=center,
                minimum height=2.7cm},
  blkorch/.style={rectangle, rounded corners=6pt, draw=grpC, thick,
                  fill=grpC!9, text width=4.0cm, align=center,
                  minimum height=3.3cm},
  iobox/.style={rectangle, rounded corners=4pt, draw=stagegray,
                fill=stagegray!18, font=\small, align=center,
                minimum height=0.75cm, text width=2.1cm},
  kgbox/.style={rectangle, rounded corners=4pt, draw=grpA, dashed,
                fill=grpA!12, font=\small\bfseries, text=grpA,
                align=center, minimum width=3.9cm, minimum height=0.65cm},
  scale=0.85, transform shape]

  % Three functional blocks
  \node[blkllm] (fllm) {%
    {\color{stageblue}\bfseries\normalsize $f_{\!\mathrm{llm}}$}\\[2pt]
    \textcolor{stageblue}{\rule{\linewidth}{0.5pt}}\\[2pt]
    {\footnotesize
      \textbf{S0}\enspace Necessity check\\[2pt]
      \textbf{S1}\enspace Soft classify\\[2pt]
      \textbf{S2}\enspace Slot extraction}%
  };

  \node[blkkg, right=of fllm] (fkg) {%
    {\color{grpA}\bfseries\normalsize $f_{\!\mathcal{K}}$}\\[2pt]
    \textcolor{grpA}{\rule{\linewidth}{0.5pt}}\\[2pt]
    {\footnotesize
      \textbf{S3}\enspace KG-driven elicitation loop:\\[2pt]
      completeness\\consistency\\class revision}%
  };

  \node[blkorch, right=of fkg] (forch) {%
    {\color{grpC}\bfseries\normalsize $f_{\!\mathrm{orch}}$}\\[2pt]
    \textcolor{grpC}{\rule{\linewidth}{0.5pt}}\\[2pt]
    {\footnotesize
      \textbf{S4}\enspace User confirmation\\[2pt]
      \textbf{S5}\enspace Constrained service\\profile generation\\[2pt]
      \textbf{S6}\enspace Deployment mode /\\fallback assignment}%
  };

  % Shared y-coordinate for all three user-side nodes
  \coordinate (sharedY) at ([yshift=1.0cm]forch.north);

  % NL service intent
  \node[iobox, anchor=center] (inp) at (sharedY -| fllm)
    {$\mathbf{i}$\\[-1pt]\footnotesize NL service intent};

  % User clarification response
  \node[rectangle, rounded corners=4pt, draw=grpA, dashed, fill=grpA!9,
        font=\footnotesize\bfseries, text=grpA, align=center, text width=2.2cm,
        anchor=center] (userresp) at (sharedY -| fkg)
    {User clarification\\response};

  % Orchestration-ready service profile
  \node[iobox, draw=stageblue!70, fill=stageblue!12, text width=2.2cm,
        anchor=center] (out) at (sharedY -| forch)
    {$\mathcal{P}$\\[-1pt]\footnotesize orchestration-ready\\service profile};

  % Knowledge graph
  \node[kgbox, below=0.95cm of fkg] (kg)
    {Knowledge Graph $\mathcal{K}$};
  \draw[thick, {Stealth[length=4pt]}-{Stealth[length=4pt]}, grpA]
    (fkg.south) -- (kg.north);
  \draw[dashed, thick, -{Stealth[length=4pt]}, grpA]
    (kg.west) -| (fllm.south)
    node[near start, above, font=\scriptsize, text=grpA] {taxonomy};

  % TN/NTN context
  \node[rectangle, rounded corners=4pt, draw=grpC, dashed, fill=grpC!9,
        font=\footnotesize, text=grpC, align=left, text width=5.0cm,
        below=0.35cm of forch] (tnctx) {%
    \textcolor{grpC}{\bfseries Integrated TN/NTN Network Context}\\[2pt]
    \textbullet~TN availability\\
    \textbullet~NTN availability\\
    \textbullet~Scenario / coverage state\\
    \textbullet~Segment constraints%
  };
  \draw[thick, -{Stealth[length=4pt]}, grpC]
    (tnctx.north) -- (forch.south);

  % Clarification arrows
  \draw[thick, -{Stealth[length=4pt]}, grpA]
    ([xshift=-4pt]fkg.north) --
    node[left, font=\scriptsize, text=grpA, align=right] {clarification\\query}
    ([xshift=-4pt]userresp.south);
  \draw[thick, -{Stealth[length=4pt]}, grpA]
    ([xshift=4pt]userresp.south) --
    node[right, font=\scriptsize, text=grpA, align=left] {user\\response}
    ([xshift=4pt]fkg.north);

  % Main flow
  \draw[pipe] (inp)   -- (fllm);
  \draw[pipe] (fllm)  -- (fkg);
  \draw[pipe] (fkg)   -- (forch);
  \draw[pipe] (forch) -- (out);

  % Fast path
  \draw[pipe, dashed, bend right=24]
    (fllm.south east) to
    node[left=0.7cm, font=\scriptsize, midway, align=center]
    {fast-path}
    (forch.south west);

  % Feedback path
  \draw[thick, -{Stealth[length=5pt]}, stagegray, bend right=18]
    (forch.north west) to
    node[above, font=\scriptsize, midway, yshift=2pt] {user corrects}
    (fkg.north east);

  % Left-side braces
  \coordinate (braceX) at ([xshift=-0.9cm]fllm.west);

  % User-side brace
  \draw[decorate, decoration={brace, amplitude=4pt},
        thick, stagegray]
    (braceX |- inp.south) -- (braceX |- inp.north)
    node[midway, left=7pt, font=\scriptsize\bfseries, text=stagegray, align=right]
    {USER};

  % System-side brace
  \draw[decorate, decoration={brace, amplitude=4pt},
        thick, stagegray]
    ([yshift=-1.35cm]braceX |- forch.south) -- (braceX |- forch.north)
    node[midway, left=7pt, font=\scriptsize\bfseries, text=stagegray, align=right]
    {SYSTEM};

\end{tikzpicture}
\caption{Proposed framework architecture implementing the three-stage intent-to-profile mapping introduced in Section~\ref{sec:system_model}. The fast-path (dashed) bypasses $f_{\mathcal{K}}$ when all required intent slots are extractable from $\mathbf{i}$ without dialogue. The feedback arc returns to $f_{\mathcal{K}}$ when the user corrects the Stage~4 confirmation. The integrated TN/NTN network context feeds into $f_{\mathrm{orch}}$ to support deployment mode and fallback assignment at Stage~6.}
\vspace{-1em}
\label{fig:pipeline}
\end{figure*}

\subsection{Integrated TN/NTN Context}
In addition to service semantics, the orchestration process depends on the TN/NTN context state $\mathbf{z} \in \mathcal{Z}$, the set of admissible integrated network conditions, including high-level attributes such as TN availability, NTN availability, scenario type, coverage condition, and segment-specific constraints.

We consider four representative TN/NTN environments:
\begin{itemize}[itemsep=1pt,topsep=2pt]
    \item \textbf{Urban:} strong terrestrial support, dense users, and latency-sensitive services;
    \item \textbf{Rural/Remote:} sparse terrestrial coverage with strong continuity requirements;
    \item \textbf{Disaster/Emergency:} degraded terrestrial infrastructure where resilience and fallback are critical;
    \item \textbf{Maritime:} limited or absent terrestrial support with NTN-dominant connectivity.
\end{itemize}

\subsection{Knowledge-Graph-Guided Intent-to-Profile Mapping}
The framework maps the input intent to the final service profile through
\begin{equation}
    \mathcal{P} = f_{\mathrm{orch}}\!\left(f_{\mathcal{K}}\!\left(f_{\mathrm{llm}}(\mathbf{i})\right)\right),
    \label{eq:overall_mapping}
\end{equation}
where $f_{\mathrm{llm}}(\cdot)$ performs semantic interpretation and initial field extraction, $f_{\mathcal{K}}(\cdot)$ applies knowledge-guided validation and clarification, and $f_{\mathrm{orch}}(\cdot)$ generates the final orchestration-ready profile. In the proposed framework, $f_{\mathrm{orch}}(\cdot)$ also incorporates the TN/NTN context $\mathbf{z}$ to produce a deployment-aware service profile.

\subsection{Problem Formulation}

Given a natural-language service request $\mathbf{i}$ and an integrated TN/NTN context state $\mathbf{z}$, the goal is to determine a valid orchestration-ready profile
\begin{equation}
    \mathcal{P}^{*} \in \mathcal{P}_{\mathrm{valid}},
    \label{eq:problem}
\end{equation}
that correctly interprets $\mathbf{i}$, resolves missing or ambiguous requirements through clarification, detects and handles contradictions, satisfies the service ontology and domain constraints, and assigns a deployment mode consistent with $\mathbf{z}$. This problem is addressed at the higher-layer intent-to-service-profile interface rather than at the physical or radio-resource-control level, focusing on semantic interpretation, knowledge-guided validation, and profile generation ahead of downstream TN/NTN orchestration. The resulting profiles are evaluated with the intent-level metrics of Section~\ref{sec:evaluation}: classification accuracy, policy validity, contradiction handling, and clarification efficiency.

\section{KNOWLEDGE-GRAPH GUIDED INTENT PROCESSING}
\label{sec:framework}

\subsection{Pipeline Overview}
As shown in Fig.~\ref{fig:pipeline}, the proposed pipeline maps a natural-language intent $\mathbf{i}$ to a validated service profile $\mathcal{P}$ through three coupled modules: an LLM-based semantic parser, a knowledge-guided validation layer, and a final orchestration layer. The knowledge graph $\mathcal{K}$ provides the service taxonomy, slot definitions, conflict rules, class-revision rules, and deployment constraints used throughout the pipeline; the approach is similar to \cite{mehmood2024KG}.

$\mathcal{K}$ is manually curated around the four service classes of Section~\ref{sec:system_model} and encodes five typed relations: \textsc{requires\_parameter} (mandatory slots for a class), \textsc{requires\_conditional} (slots that become mandatory only once a prior answer triggers them), \textsc{similar\_to} (a soft class-revision proposal when elicited attributes suggest a neighboring class), \textsc{escalates\_to} (a stronger class transition when the requested KPI envelope exceeds the current class), and \textsc{conflict\_rule} (pairwise incompatible slot values, e.g., a Critical priority with a LowCost profile, or a GEO-satellite tier with sub-100~ms latency). Extending $\mathcal{K}$ to additional service classes or finer-grained KPI models requires adding new slot and conflict-rule entries under the same five relation types, without changes to the pipeline logic in Fig.~\ref{fig:pipeline}.

\subsection{LLM-Based Semantic Parsing}
The LLM module performs two tasks; soft class inference and slot extraction. Given $\mathbf{i}$, it first predicts a provisional service class
\begin{equation}
    c = f_{\mathrm{cls}}(\theta;\mathbf{i}),
    \label{eq:stage1}
\end{equation}
where $\theta$ denotes the LLM parameters. This classification is treated as a
soft hypothesis: as the KG elicitation dialogue progresses and the user resolves
ambiguities, the accumulated slot values may reveal a better-fitting class.
The KG layer can then revise $c$ via a \textsc{similar\_to} relation (when the
collected attributes suggest a neighbouring class) or escalate it via
\textsc{escalates\_to} (when the requested KPI envelope exceeds the capacity of
the current class). Any revision is proposed explicitly to the user and never
applied silently, ensuring the final class assignment reflects a confirmed
interpretation rather than a unilateral model decision. Conditioned on $c$, the model then extracts a partial slot representation
\begin{equation}
    \hat{\mathbf{s}} = f_{\mathrm{gen}}(\mathbf{i},c;\theta).
    \label{eq:stage2}
\end{equation}
Both stages use grammar-constrained decoding so that outputs remain schema-valid. The extracted slots are compared against the mandatory field set $\mathcal{F}_c$ defined by $\mathcal{K}$. If all required slots are already available, the pipeline follows the fast path and skips clarification. Otherwise, unresolved fields are passed to the KG layer:
\begin{equation}
    \mathcal{M}=\{f\in\mathcal{F}_c \mid \hat{s}_f=\emptyset\}.
    \label{eq:fastpath}
\end{equation}

\subsection{KG-Guided Validation and Clarification}

The KG layer is responsible for completeness checking, contradiction detection, and class revision. During this stage, the unresolved set is updated dynamically as conditional rules are activated:
\begin{equation}
\mathcal{M}
=
\left\{
f \in \mathcal{F}
\,\middle|\,
\hat{s}_f = \emptyset
\ \text{or}\
\hat{s}_f \notin \mathcal{V}_f
\right\},
\label{eq:missing}
\end{equation}
where $\mathcal{F}$ is the current mandatory field set and $\mathcal{V}_f$ is the valid-value domain of slot $f$.

Consistency is enforced through conflict rules encoded in $\mathcal{K}$:
\begin{equation}
    \exists (r_i,r_j)\in\hat{\mathbf{s}}\times\hat{\mathbf{s}}
    : (r_i,r_j)\in\mathcal{R}_{\mathrm{conflict}},
    \label{eq:conflict}
\end{equation}
where $\mathcal{R}_{\mathrm{conflict}}$ denotes the set of incompatible slot-value pairs. If missing, low-confidence, or conflicting attributes remain, the framework opens a clarification loop and queries the user only for the unresolved items. This loop terminates once the representation is complete, conflict-free, and class-consistent. The resulting output is the validated intent $\mathbf{s}^{*}$.

\subsection{Profile Finalization}

The final module converts $\mathbf{s}^{*}$ into the orchestration-ready profile $\mathcal{P}$. First, the framework summarizes the interpreted request and allows user confirmation or correction. It then emits the final structured profile using constrained generation. Finally, the deployment mode is assigned in a domain-grounded manner: if a user preference is explicitly provided and remains valid after KG checking, it is retained; otherwise, the mode is selected from the deployment rules in $\mathcal{K}$ using the validated service class and scenario context.

Accordingly, the output profile contains the resolved service class, deployment mode, fallback policy, and the validated service attributes required by downstream orchestration functions. The profile $\mathcal{P}$ therefore serves as the interface between natural-language service intent and structured TN/NTN service provisioning.

%%%%%%%%%%%%%%%%%%%%%%%%%%%%%%%%%%%%%%%%%%%%%%%%%%%%%%%%%%%%%%%%%%%%%%%%%%%%%%%%
% === V. PERFORMANCE EVALUATION
%%%%%%%%%%%%%%%%%%%%%%%%%%%%%%%%%%%%%%%%%%%%%%%%%%%%%%%%%%%%%%%%%%%%%%%%%%%%%%%%
\section{PERFORMANCE EVALUATION}
\label{sec:evaluation}

This section evaluates the proposed knowledge-guided LLM framework at the intent level, focusing on whether it correctly interprets natural-language service requests and produces valid orchestration-ready profiles.

\subsection{Dataset}
\label{sec:dataset}
\begin{figure}[t]
\centering
\begin{tikzpicture}
\begin{axis}[
    ybar=0pt,
    bar width=10pt,
    width=\linewidth,
    height=4.6cm,
    ymin=0, ymax=100,
    ylabel={Score (\%)},
    symbolic x coords={Acls,Tbar,Vpol,Aconf},
    xtick=data,
    xticklabels={
        {$A_{\mathrm{cls}}$},
        {$\bar{T}$},
        {$V_{\mathrm{pol}}$},
        {$A_{\mathrm{conf}}$}
    },
    enlarge x limits=0.22,
    ymajorgrids=true,
    grid style={dashed,gray!30},
    axis line style={black!60},
    tick style={black!60},
    ylabel style={font=\small},
    tick label style={font=\small},
    xticklabel style={align=center},
    legend style={
        at={(0.98,0.98)},
        anchor=north east,
        font=\scriptsize,
        draw=gray!50,
        fill=white,
        rounded corners=2pt,
        legend cell align=left
    },
    legend image post style={scale=1.0},
]

% KG-guided
\addplot[
    fill=stageblue!45,
    draw=stageblue,
    line width=0.6pt
] coordinates {
    (Acls,74)
    (Tbar,18)
    (Vpol,96.4)
    (Aconf,61.4)
};

% Pure-LLM
\addplot[
    fill=grpA!35,
    draw=grpA,
    line width=0.6pt
] coordinates {
    (Acls,74.4)
    (Tbar,62.3)
    (Vpol,76.8)
    (Aconf,26.4)
};

\legend{KG-guided, Pure-LLM}
\end{axis}
\end{tikzpicture}
\vspace{-0.4em}
\caption{KG-guided vs.\ Pure-LLM reasoning on the 100-sample contradictory-only subset ($N=100$, Section~\ref{sec:dataset}). Metrics are reported in percentage except $\bar{T}$ (average clarification turns).}
\label{fig:pure_llm_bar}
\end{figure}
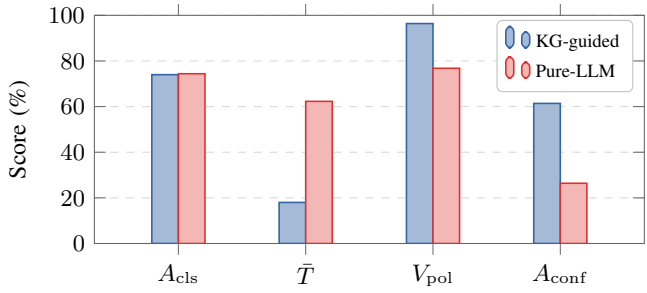

We construct a 200-sample labeled intent dataset using a combination of hand-authored seed intents and LLM-assisted variants, balanced across the four service classes of Section~\ref{sec:system_model} (50 samples per class). For each class, five seed intents are hand-authored from realistic vertical scenarios (e.g., industrial robotic control, public-safety fallback, maritime/rural NTN reach, city-scale IoT metering); each seed is then expanded into paraphrased variants (reworded, same requirements), ambiguous variants (one or more mandatory slots deliberately omitted to exercise the KG elicitation loop), and contradictory variants (a documented conflict-rule instance, e.g., Critical priority with a LowCost profile, or GEO-satellite NTN with sub-100~ms latency). Every sample carries an expected service class, an expected slot subset, and, for contradictory samples, the expected conflict to be flagged. Paraphrased and ambiguous text is LLM-assisted for lexical variety, but every ground-truth label is authored and manually verified against the domain ontology by the authors rather than taken from LLM output. Since the LLM stages operate zero-shot under grammar-constrained decoding with no fine-tuning on this data, the dataset is used purely for evaluation and no train/test split is required. A separate 100-sample contradictory-only subset, spanning all eight documented conflict-rule types, is used for the KG rule-coverage sweep (Fig.~\ref{fig:kg_sweep}) and the Pure-LLM comparison (Fig.~\ref{fig:pure_llm_bar}) to isolate conflict-detection behavior from the completeness behavior of well-formed seed intents.

\subsection{Performance Metrics}
We use four performance metrics, intent classification accuracy $A_{\mathrm{cls}}$, which measures correct class prediction; policy validity rate $V_{\mathrm{pol}}$, which measures the fraction of generated profiles satisfying schema and domain constraints; contradiction detection accuracy $A_{\mathrm{conf}}$, which measures how reliably inconsistent requests are identified; and clarification efficiency $\bar{T}$, which measures the average number of elicitation turns. Lower $\bar{T}$ indicates that more information is extracted directly from the input without iterative clarification. Deployment mode is not evaluated separately, since in the current framework it is assigned deterministically after profile validation.

\subsection{Results and Discussion}

Table~\ref{tab:results_intent} reports the intent-level results on the 200-sample dataset described in Section~\ref{sec:dataset}. Overall, the proposed framework achieves \(A_{\mathrm{cls}}=89\%\), \(V_{\mathrm{pol}}=95\%\), and \(\bar{T}=0.66\) turns per intent, indicating that most requests are correctly interpreted and finalized with limited clarification overhead.

\begin{table}[t]
\centering
\caption{Intent-level evaluation results per service class ($n$=200).}
\label{tab:results_intent}
\footnotesize
\renewcommand{\arraystretch}{1.02}
\setlength{\tabcolsep}{6pt}
\begin{tabular}{c|c|c|c|c}
\toprule
\textbf{Class} & $n$ & $A_{\mathrm{cls}}$ & $V_{\mathrm{pol}}$ & $\bar{T}$ \\
\midrule
TC  & 50 & 98\% &  83\% & 0.92 \\
MC  & 50 & 94\% & 98\% & 0.26 \\
RC  & 50 & 97\% & 98\% & 0.64 \\
MaC & 50 & 67\% & 100\% & 0.83 \\
\midrule
\textbf{All} & \textbf{200} & \textbf{89\%} & \textbf{95\%} & \textbf{0.66} \\
\bottomrule
\end{tabular}
\end{table}

At the class level, TC and RC achieve high classification accuracies of \(98\%\) and \(97\%\), respectively, while MC remains high at \(94\%\). The main source of classification error is MaC (\(67\%\)), where wide-area telemetry and IoT-oriented requests partially overlap with RC in their coverage-driven language, indicating that the dominant classification challenge lies in the RC--MaC boundary rather than in the overall pipeline design. MaC achieves \(V_{\mathrm{pol}}=100\%\), while MC and RC both achieve \(98\%\), showing that once the class is resolved, the KG layer reliably enforces profile completeness and rule consistency. TC is the only class with reduced validity (\(83\%\)), mainly because it depends on stricter control-oriented fields such as \texttt{application\_type} and \texttt{cycle\_time\_ms}, which are not always recoverable within the turn budget.

Clarification efficiency further highlights the role of the KG layer. MC is handled predominantly through the fast path with \(\bar{T}=0.26\), whereas TC and MaC require more interaction due to tighter constraints and higher semantic ambiguity. The overall value \(\bar{T}=0.66\) confirms that most directly available attributes are extracted without iterative dialogue, while the KG loop is invoked primarily for unresolved or conflicting slots rather than routine field filling.

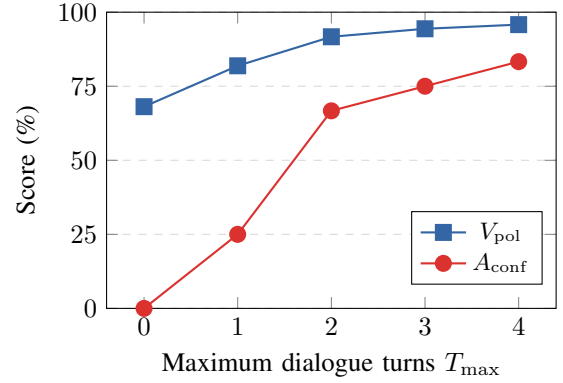
\begin{figure}[t]
\centering
\begin{tikzpicture}
\begin{axis}[
  width=0.85\columnwidth,
  height=5.5cm,
  xlabel={Maximum dialogue turns $T_{\max}$},
  ylabel={Score (\%)},
  xtick={0,1,2,3,4},
  xticklabels={$0$,$1$,$2$,$3$,$4$},
  ymin=0, ymax=100,
  ytick={0,25,50,75,100},
  ymajorgrids=true,
  grid style={dashed,gray!30},
  legend style={at={(0.97,0.08)}, anchor=south east, font=\small},
  every axis plot/.append style={thick, mark size=2.8pt},
]
\addplot[color=stageblue, mark=square*]
  coordinates {(0,68.1) (1,81.9) (2,91.7) (3,94.4) (4,95.8)};
\addplot[color=grpA, mark=*]
  coordinates {(0,0.0) (1,25.0) (2,66.7) (3,75.0) (4,83.3)};
\legend{$V_{\mathrm{pol}}$, $A_{\mathrm{conf}}$}
\end{axis}
\end{tikzpicture}
\caption{Turn-budget sensitivity: $V_{\mathrm{pol}}$ and $A_{\mathrm{conf}}$
vs.\ maximum KG dialogue turns $T_{\max}$ ($N=200$, $T_{\max}=4$ is the
actual eval baseline). $V_{\mathrm{pol}}$ starts at $68.1\%$ even without
dialogue and saturates quickly. $A_{\mathrm{conf}}$ starts at $0\%$: detecting
contradictions requires the KG loop to elicit both conflicting slots before the
conflict rule can fire.}
\label{fig:turn_budget}
\end{figure}

Fig.~\ref{fig:turn_budget} shows the sensitivity of profile validity and contradiction handling to the dialogue budget \(T_{\max}\). Even at \(T_{\max}=0\), the framework attains \(V_{\mathrm{pol}}=68.1\%\), indicating that a substantial fraction of profiles can be completed directly from the initial intent. With only two clarification turns, \(V_{\mathrm{pol}}\) rises to \(91.7\%\), already approaching the \(T_{\max}=4\) baseline result. In contrast, \(A_{\mathrm{conf}}\) starts at \(0\%\) and increases more gradually, since contradiction detection requires both conflicting attributes to be surfaced before a rule can fire. This result separates the roles of the two layers: semantic extraction primarily drives completeness, whereas contradiction handling is inherently dialogue-dependent.

\begin{figure}[t]
\centering
\begin{tikzpicture}
\begin{axis}[
  width=0.95\columnwidth,
  height=4.5cm,
  xlabel={KG conflict-rule coverage (\%)},
  ylabel={Score (\%)},
  xtick={0,25,50,75,100},
  ymin=0, ymax=100,
  ytick={0,25,50,75,100},
  ymajorgrids=true,
  grid style={dashed,gray!30},
  legend style={at={(0.05,0.97)}, anchor=north west, font=\small},
  every axis plot/.append style={thick, mark size=2.8pt},
]
\addplot[color=grpA, mark=*]
  coordinates {(0,0.0) (25,32.5) (50,35.0) (75,62.5) (100,85.7)};
\addplot[color=grpC, mark=triangle*]
  coordinates {(0,0.0) (25,28.7) (50,30.0) (75,51.2) (100,71.4)};
\legend{$A_{\mathrm{conf}}$, $R_{\mathrm{cf}}$}
\end{axis}
\end{tikzpicture}
\caption{KG conflict-rule coverage sweep on the 100-sample contradictory-only subset ($N=100$, Section~\ref{sec:dataset}):
$A_{\mathrm{conf}}$ (contradiction detection accuracy) and $R_{\mathrm{cf}}$
(conflict detection rate) as the fraction of active KG rules increases from
$0\%$ to $100\%$.  Both curves are step-shaped — each plateau corresponds to
rules that do not apply to the test sample set; each jump corresponds to a new
class of rule (universal priority/cost at $25\%$; deployment-scenario rules at
$75\%$; full class-specific rules at $100\%$) becoming active.}
\label{fig:kg_sweep}
\end{figure}
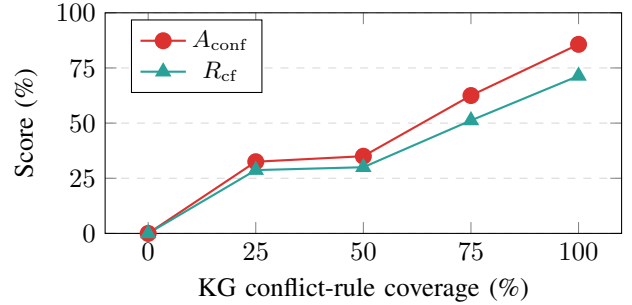

Fig.~\ref{fig:kg_sweep} isolates the effect of KG rule coverage. As the active rule set increases from \(0\%\) to \(100\%\), both contradiction detection accuracy and conflict detection rate improve with the largest gain obtained when deployment-scenario rules become active. The step-shaped behavior reflects the fact that different rule groups apply to different subsets of the contradiction-focused evaluation set. The final point confirms that the KG contributes not only to completeness but also to explicit contradiction awareness, especially for cases that cannot be resolved reliably through unconstrained text generation alone.

As Table~\ref{tab:feature_comparison} shows, no prior work performs the same end-to-end mapping to a validated, deployment-aware TN/NTN profile, so no external baseline is directly comparable; we instead isolate the KG layer's contribution via a Pure-LLM ablation, replacing Stage~3 with an LLM-driven loop that determines completeness, missing fields, and conflicts from parametric knowledge and prompt-embedded class descriptions alone, while keeping all other stages unchanged. Since both systems share the Stage-1 classifier, the nearly identical \(A_{\mathrm{cls}}\) reflects a common front-end; the differences emerge downstream, where the KG-guided system improves \(V_{\mathrm{pol}}\) by \(19.6\) percentage points and \(A_{\mathrm{conf}}\) by \(35\) percentage points while using approximately \(3.5\times\) fewer clarification turns. The KG's benefit is thus controlled elicitation, deterministic conflict checking, and reliable finalization, not class prediction: the Pure-LLM loop asks more questions yet still yields more incomplete or inconsistent profiles.

\section{Conclusion}\label{sec:conclusion}
This paper proposed a knowledge-guided LLM framework for intent-driven service orchestration in integrated TN/NTN 6G networks. By combining grammar-constrained LLM extraction with a KG-driven elicitation loop, the framework converts natural-language service requests into validated, orchestration-ready service profiles, and the KG layer consistently outperforms a Pure-LLM baseline in profile validity, contradiction detection, and clarification efficiency. Evaluation on the expanded 200-sample intent dataset further demonstrates that the framework maintains high profile validity with limited clarification overhead. The present study serves as a proof of concept: the knowledge graph is intentionally kept compact to capture the dominant service classes and conflict patterns in integrated TN/NTN settings, and can be extended with richer ontologies, finer-grained KPI models, and additional multi-segment conflict rules to support more complex service scenarios. Future work will focus on closing the full intent-to-orchestration loop by using the generated service profiles as input to downstream TN/NTN resource management and incorporating feedback from service outcomes into the intent-processing layer.

\bibliographystyle{IEEEtran}

\end{document}